\documentclass[acmsmall]{acmart}
\usepackage{epsfig}
\usepackage{multirow}
\usepackage{graphicx}
\usepackage{amsmath}
\usepackage{amssymb}
\usepackage{amsmath}

\usepackage{algorithm} 
\usepackage{algpseudocode}  
\AtBeginDocument{%
  \providecommand\BibTeX{{%
    \normalfont B\kern-0.5em{\scshape i\kern-0.25em b}\kern-0.8em\TeX}}}

\begin{document}

%%
%% The "title" command has an optional parameter,
%% allowing the author to define a "short title" to be used in page headers.
\title{Two-Level Residual Distillation based Triple Network for Incremental Object Detection}

\author{Dongbao Yang}
\affiliation{\institution{Institute of Information Engineering, Chinese Academy of Sciences}}
\email{yangdongbao@iie.ac.cn}

\author{Yu Zhou}
\affiliation{\institution{Institute of Information Engineering, Chinese Academy of Sciences}}
\email{zhouyu@iie.ac.cn}

\author{Dayan Wu}
\affiliation{\institution{Institute of Information Engineering, Chinese Academy of Sciences}}
\email{wudayan@iie.ac.cn}

\author{Can Ma}
\affiliation{\institution{Institute of Information Engineering, Chinese Academy of Sciences}}
\email{macan@iie.ac.cn}
\author{Fei Yang}
\affiliation{\institution{TAL Education Group}}
\email{yang.fei@100tal.com}
\author{Weiping Wang}
\affiliation{\institution{Institute of Information Engineering, Chinese Academy of Sciences}}
\email{wangweiping@iie.ac.cn}
%%
%% By default, the full list of authors will be used in the page
%% headers. Often, this list is too long, and will overlap
%% other information printed in the page headers. This command allows
%% the author to define a more concise list
%% of authors' names for this purpose.
\renewcommand{\shortauthors}{D. Yang, et al.}

%%
%% The abstract is a short summary of the work to be presented in the
%% article.
\begin{abstract}
  Modern object detection methods based on convolutional neural network suffer from severe catastrophic forgetting in learning new classes without original data. Due to time consumption, storage burden and privacy of old data, it is inadvisable to train the model from scratch with both old and new data when new object classes emerge after the model trained. In this paper, we propose a novel incremental object detector based on Faster R-CNN to continuously learn from new object classes without using old data. It is a triple network where an old model and a residual model as assistants for helping the incremental model learning on new classes without forgetting the previous learned knowledge. To better maintain the discrimination of features between old and new classes, the residual model is jointly trained on new classes in the incremental learning procedure. In addition, a corresponding distillation scheme is designed to guide the training process, which consists of a two-level residual distillation loss and a joint classification distillation loss. Extensive experiments on VOC2007 and COCO are conducted, and the results demonstrate that the proposed method can effectively learn to incrementally detect objects of new classes, and the problem of catastrophic forgetting is mitigated in this context.
\end{abstract}

%%
%% The code below is generated by the tool at http://dl.acm.org/ccs.cfm.
%% Please copy and paste the code instead of the example below.
%%


%%
%% Keywords. The author(s) should pick words that accurately describe
%% the work being presented. Separate the keywords with commas.
%\begin{CCSXML}
%<ccs2012>
% <concept>
%  <concept_id>10010520.10010553.10010562</concept_id>
%  <concept_desc>Computer systems organization~Embedded systems</concept_desc>
%  <concept_significance>500</concept_significance>
% </concept>
% <concept>
%  <concept_id>10010520.10010575.10010755</concept_id>
%  <concept_desc>Computer systems organization~Redundancy</concept_desc>
%  <concept_significance>300</concept_significance>
% </concept>
% <concept>
%  <concept_id>10010520.10010553.10010554</concept_id>
%  <concept_desc>Computer systems organization~Robotics</concept_desc>
%  <concept_significance>100</concept_significance>
% </concept>
% <concept>
%  <concept_id>10003033.10003083.10003095</concept_id>
%  <concept_desc>Networks~Network reliability</concept_desc>
%  <concept_significance>100</concept_significance>
% </concept>
%</ccs2012>
%\end{CCSXML}
%
%\ccsdesc[500]{Computer systems organization~Embedded systems}
%\ccsdesc[300]{Computer systems organization~Redundancy}
%\ccsdesc{Computer systems organization~Robotics}
%\ccsdesc[100]{Networks~Network reliability}
%
%%%
%%% Keywords. The author(s) should pick words that accurately describe
%%% the work being presented. Separate the keywords with commas.
%\keywords{datasets, neural networks, gaze detection, text tagging}

%%
%% This command processes the author and affiliation and title
%% information and builds the first part of the formatted document.
\maketitle
\section{Introduction}
\begin{figure}[t]
	\begin{center}
		%\fbox{\rule{0pt}{2in} \rule{0.9\linewidth}{0pt}}
		\includegraphics[width=0.7\linewidth]{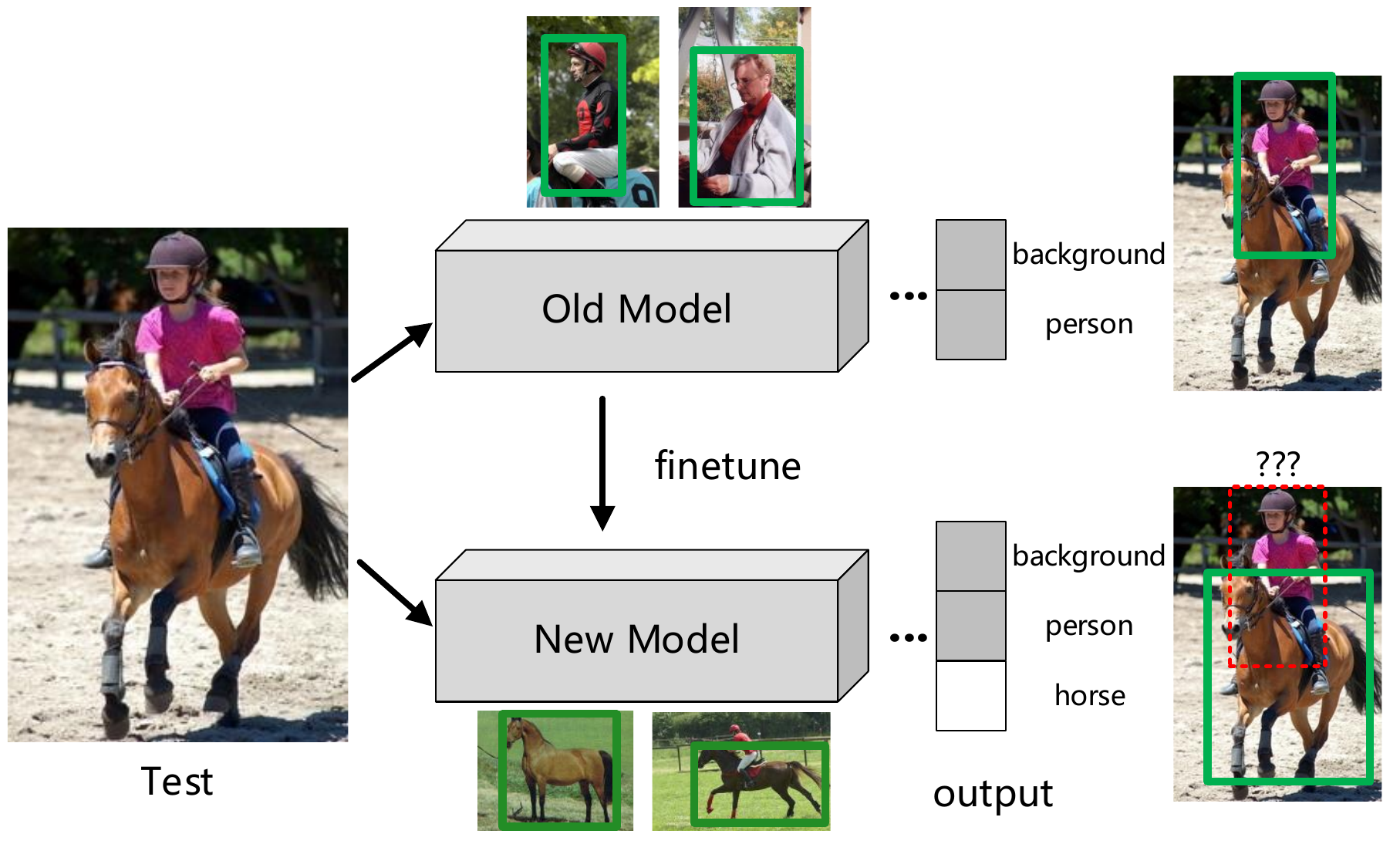}
	\end{center}
	\caption{Finetuning on new classes. The old model is trained on old class (person), and new model is trained by finetuning old model on new class (horse) by adding one unit on the output layer. The person in the test image can not be detected after finetuning.}
	\label{fig:ft}
\end{figure}

Despite modern object detection methods based on convolutional neural network have achieved state-of-the-art results, it suffers from severe catastrophic forgetting~\cite{french1999catastrophic}~\cite{goodfellow2013empirical}~\cite{mccloskey1989catastrophic} in learning new object classes. In practice, new object classes often emerge after the model trained. 
%Finetuning is a common way to transfer the pretrained model to new data or tasks. However, if the model is finetuned with only the data of previously unseen classes, it would completely forget the old classes~\cite{kirkpatrick2017overcoming}, as shown in Figure~\ref{fig:ft}, which is the result of finetuning model on one new class (tvmonitor). It can be seen that the ability to detect old classes (cat, person, table and etc.) is severely degraded. 
Finetuning is a common way to adapt the old model to new classes, which is achieved by replacing the output layer with new classes or by adding units in the output layer for new classes, as shown in Figure~\ref{fig:ft}. However, the performance may degrade severely due to the absence of old data.
Intuitively, training the model from scratch with both old and new data would solve this problem, but it will take a lot of time and increase the storage burden for storing old data. In particular, the training data for the pretrained model are not always available for a new task. Therefore, it is necessary to develop incremental learning methods for object detection, which can continuously learn from new data instead of training on the whole dataset and preserve the previously learned knowledge.

Many studies on incremental learning mainly focus on image classification. According to the optimization directions to overcome catastrophic forgetting, these methods can be divided into two categories~\cite{hou2019learning}: (1) preserving significant parameters of the original model~\cite{aljundi2018memory}~\cite{kirkpatrick2017overcoming}~\cite{zenke2017continual}; (2) preserving the knowledge of the original model through knowledge distillation~\cite{aljundi2017expert}~\cite{jung2018less}~\cite{li2017learning}~\cite{rannen2017encoder}~\cite{rebuffi2017icarl}. Due to the fact that it is difficult to design a reasonable metric to evaluate the importance of all parameters, we follow the second direction to preserve the knowledge of the original classes when adapting the model to detect new object classes, which utilizes the supervised information provided by old model to guide the training of new model by distillation losses. 
%In order to maintain the performance on the old tasks, some studies keep an exemplar set from old data, and use both these exemplars and the new data for training~\cite{castro2018end}~\cite{chaudhry2018riemannian}~\cite{hou2018lifelong}~\cite{lopez2017gradient}~\cite{rebuffi2017icarl}. %The exemplars are chosen at random or according to some metrics.
%In ~\cite{castro2018end}, representative memory unit performs new samples selection based on herding selection, which produces a sorted list of samples of one class based on the distance to the mean sample, and the first $n$ samples are selected.~\cite{rebuffi2017icarl} also proposes a method to select a small number of exemplars from old classes.
%However, the original training data may not be available for some special cases, such as privacy issues or limited storage budgets. 

Different from image classification, object detection involves the discrimination between foreground and background and the precise localization of objects, which increases the difficulty of incremental learning. Previous incremental object detection methods~\cite{chen2019new}~\cite{hao2019take}~\cite{hao2019end}~\cite{li2019rilod}~\cite{shmelkov2017incremental}~\cite{zhang2020class} mainly directly let the incremental model imitate the old model in some important positions such as the feature space and the output layers to preserve the learned knowledge from old data, which is achieved by constraining the activations between them to be similar. However, simply imitating old model will suppress the feature discrimination between old and new classes to some extent. 

To solve these problems, we propose a novel triple network based incremental object detector, which is based on Faster R-CNN~\cite{ren2015faster}. Three detection models cooperate for adapting the network trained on old classes to new classes, and ensuring that the performance on the old classes will not degrade without using original training data. To train an incremental model which is responsible for detecting both old and new classes, a frozen copy of the original Faster R-CNN trained on old classes is utilized to provide the knowledge of old classes including feature, distributions of output layers and pseudo ground-truth. In addition, a novel residual model is proposed to assist the incremental learning procedure. To preserve the previous learned knowledge, a novel distillation scheme is designed which includes a two-level residual distillation loss and a joint classification distillation loss applied on the feature space and the output layers separately. 

The contributions are as follows:
\begin{itemize}
	\setlength{\itemsep}{0pt}
	\setlength{\parsep}{0pt}
	\setlength{\parskip}{0pt} 
	\item We propose a triple-network based incremental object detector, in which a residual model is jointly trained for detecting new classes in the incremental learning procedure, and it is introduced to fit the difference between the incremental model and the old model. %To maintain the discrimination between the features of both old and new classes.
	\item A two-level residual distillation loss is designed to maintain the feature discrimination between old and new classes, and a joint classification distillation loss is used to maintain the learned knowledge from both old and new data.
	%is designed to maintain the feature discrimination between old and new classes, and we also propose a new feature fusion way for incrementally learning new knowledge. Meanwhile, a background-excluded distillation loss is designed for the classifier of the detection model. 
	
	%\item We propose an end-to-end incremental object detection method with the learnable RPN. To the best of our knowledge, this is the first work to solve the incremental object detection problem in an end-to-end way.
	%\item A multi-distillation scheme including three distillation losses is designed for different activations of the dual-network to avoid catastrophic forgetting by penalizing the differences between them.
	%A multi-distillation scheme including three distillation losses for feature space, classification and regression is designed for a dual-network to avoid the catastrophic forgetting.
	\item Extensive experiments are conducted on VOC2007~\cite{everingham2010pascal} and COCO~\cite{lin2014microsoft}, and the results demonstrate that the proposed method is effective for incremental object detection and achieves promising results compared with other methods.% Especially, it outperforms the state-of-the-art method with the mAP of 3.83\% on VOC2007 and 6.35\% on the large scale COCO dataset.
\end{itemize}

\section{Related Work}
Incremental learning is a significant problem of machine learning~\cite{cauwenberghs2001incremental}~\cite{kuzborskij2013n}~\cite{mensink2013distance}~\cite{polikar2001learn++}. Recently, with the success of deep learning, many researchers pay more attention on incremental learning of deep neural network. Most existing incremental learning methods for vision tasks mainly focus on image classification, which means to continuously update the image classifier to recognize new classes without decreasing the accuracy on previous seen classes. Existing works can be divided into two categories based on the optimization directions for preserving learned knowledge~\cite{hou2019learning}: parameter-based and distillation-based.
\begin{figure*}
	\begin{center}
		%\fbox{\rule{0pt}{2in} \rule{.9\linewidth}{0pt}}
		\includegraphics[width=1.0\linewidth]{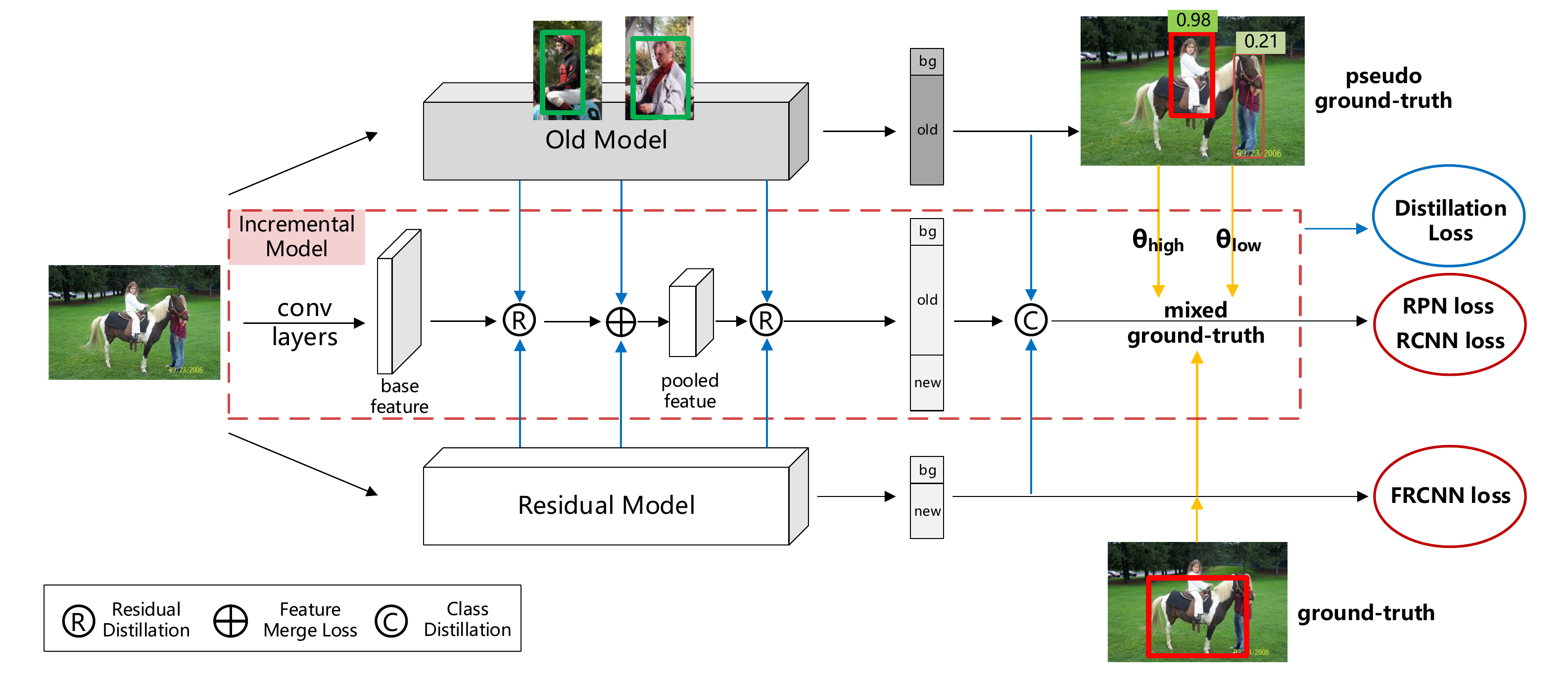}
	\end{center}
	\caption{The framework of the proposed end-to-end incremental object detector. There are three detection models: (1) Old Model is a frozen copy of original trained model, which is used to generate pseudo ground-truth and supervisory information of old classes; (2) Incremental Model is finetuned to incrementally learn new classes, while preserving the original knowledge by distillation losses; (3) Residual Model is used to learn the residual between Old Model and Incremental Model, while learning to detect new classes.}
	\label{fig:framework}
\end{figure*}

Some works are based on preserving important parameters of the network to maintain the performance on old tasks.~\cite{jung2016less} presents a method to maintain the performance on previous task by freezing the weights of the softmax layer and minimizing the distance between the features of old data extracted from target and source networks respectively. The limitation of this approach is that the weights of new and old tasks may be conflicted, and the parameters can not be updated, which will result in the degradation for learning new tasks. EWC~\cite{kirkpatrick2017overcoming} is a prominent work in this category, which sets the weights of the new model by those of the original model according to the importance of weights, and the approach remembers old tasks by selectively slowing down learning on the weights important for those tasks. MAS~\cite{aljundi2018memory} accumulates an importance measure for each parameters of the network based on the sensitivity of the predicted output function to the changes in parameters. The changes of important parameters will be penalized when learning a new task. Zenke et al.~\cite{zenke2017continual} introduce intelligent synapses to accumulate task relevant information over time, and exploit this information to rapidly store new memories without forgetting old ones. The limitation of these works is that it is hard to design a metric to evaluate the importance of all parameters.

Distillation-based method is the other representative category of incremental learning, where knowledge distillation is used to transfer knowledge from the original network to new network. Knowledge distillation is defined to utilize the supervisory information provided by teacher model to guide the training of student model to mimic the teacher model with distillation loss. LwF~\cite{li2017learning} is the first to use knowledge distillation for incremental learning, which utilizes a modified cross-entropy loss to preserve original knowledge with only examples from new task. iCaRL~\cite{rebuffi2017icarl} proposes to jointly learn feature representation and classifiers by combining representation learning and knowledge distillation, and a small set of exemplars is selected to perform nearest-mean-of-exemplars classification. Rannen et al.~\cite{rannen2017encoder} propose an auto-encoder based method to retain the knowledge from old tasks, which prevents the reconstructions of the features from changing and gives the space for features to adjust. Sun et al.~\cite{sun2018active}~\cite{sun2018lifelong} propose to maintain a lifelong dictionary, which is used to transfer knowledge to learn each new metric learning task.

Transfer learning is also related to incremental object detection, which uses the knowledge acquired from one task to help other tasks training. Finetuning is a representative paradigm of transfer learning, which is frequently used in the initialization of the backbone in object detection model with the trained CNN on ImageNet~\cite{krizhevsky2012imagenet}.~\cite{hinton2015distilling} transfers the knowledge from a large network to a small network by knowledge distillation, which encourages the responses of these two networks to be similar. However, transfer learning needs data for both old and new tasks to maintain the performance on old task, otherwise the performance will degrade severely if the old data are not available.

For incremental object detection,~\cite{shmelkov2017incremental} introduces the first incremental object detector based on Fast R-CNN~\cite{girshick2015fast} by applying knowledge distillation without using previous training data. Firstly, it uses EdgeBoxes~\cite{zitnick2014edge} and MCG~\cite{arbelaez2014multiscale} to precompute %about 2000 object 
proposals. Then, these proposals are fed into R-CNN after sampled to predict their categories. The model is trained with distillation losses applied on the outputs of final classification and regression layers to preserve the ability to recognize old classes. However, this method is not end-to-end and the proposal generation procedure is not learnable.% Due to the fact that RPN is class-sensitive, it is necessary to design specific incremental learning method for modern end-to-end two-stage object detectors.%the traditional proposal generating method is not learnable, and the process is not real-time, so that the Fast R-CNN can not be trained end-to-end and the process is not real-time too. %Due to the unlearnable character of the proposal generating process, the accuracy is affected by the quality of proposals. Besides, the long processing time results in that the Fast R-CNN is not real-time too and it can not be trained end-to-end.

Recently, several end-to-end incremental object detection methods~\cite{chen2019new}~\cite{hao2019take}~\cite{hao2019end}~\cite{li2019rilod} are proposed based on Faster RCNN~\cite{ren2015faster}.~\cite{chen2019new} proposes a hint loss to minimize difference between feature maps of the old and incremental model, and a confidence loss is used to suppress the generating of low confidence bounding boxes.~\cite{hao2019take} proposes a hierarchical large-scale retail object detection dataset (TGFS), and utilizes an exemplar set with a fixed size of old data to train an class-incremental object detector.~\cite{hao2019end} uses a frozen duplication of RPN to preserve the knowledge gained from the old classes, and a feature-changing loss is proposed to reduce the difference of the feature maps between the old and new classes.~\cite{li2019rilod} distills three types of knowledge from the old model to mimic the old model’s behavior on object classification, bounding box regression and feature extraction.~\cite{zhang2020class} proposes a dual distillation training function which pretrains a separate model only for the new classes, such that a student model can learn from two teacher models simultaneously. In addition, a novel work~\cite{perez2020incremental} proposes an incremental few-shot object detector based on CentreNet~\cite{zhou2019objects}, however, the original structure is redesigned for few-shot learning in this method. In our work, we mainly focus on the incremental object detection without changing original network, which can be applied on just giving an existing trained model and data of new classes. Different from these methods, we not only let the incremental model imitate the important activations of the old model, but also introduce a residual model trained simultaneously on new classes, which is designed to maintain the feature discrimination between old and new classes in an end-to-end way without extra model training steps. 

\section{Method}
In this paper, we bring up an end-to-end incremental object detector to continuously learn from new data without using old data. Figure~\ref{fig:framework} presents the whole framework of the proposed method. It is a triple-network which includes three detection networks. Old Model (OM) is a frozen copy of the original detector trained on old data, which provides the knowledge of old classes, including the detection results and distributions of the output layers. Incremental Model (IM) is adapted to detect both old and new classes with the annotations of new data and the knowledge from OM. The detection results from OM are regarded as the pseudo ground-truth, which are combined with the annotations of new data for updating IM. Residual Model (RM) is an assistant model jointly trained to detect new classes. To better preserve the knowledge of old classes and maintain the discrimination between old and new classes, a new distillation scheme is designed to add some constraints on the training procedure of IM, including a residual distillation loss and a joint classification distillation loss, which are applied on feature space and output layers respectively. The method is described in detail as follows.
\subsection{Triple-Network based Incremental Detector}
The triple-network for incremental object detection is based on Faster R-CNN, which is an end-to-end proposal-based object detector, and the backbone used in our framework is ResNet50~\cite{he2016deep}. %It should be noted that the structure of Old Model is the original Faster R-CNN, in other words, we can use the existing trained model to conduct incremental learning. In this paper, we train OM from scratch to better verify the performance using the split dataset. 

%For end-to-end proposal-based networks, it is difficult to directly adapt the original model to detect new classes without degrading the performance on old classes. As mentioned in~\cite{shmelkov2017incremental}, RPN in Faster R-CNN is class-sensitive. If the model is trained only on new classes, the parameters of RPN will adapt to new classes so that the ability to generate proposals corresponding to old classes will degrade. Therefore, we propose to use the detection results from OM as the pseudo ground-truth, which are combined with the ground-truth of new data to help the adaption of IM for generating proposals for both old and new classes. 

In the incremental learning stage, the parameters of IM are initialized from OM excluding the weights and bias of new added output layers for new classes, which are initialized randomly. After training samples of new data fed into the triple-network, OM will generate some bounding boxes after filtered by confidence threshold and per-category non-maxima suppression (NMS). For the remaining bounding boxes, IoU is computed between them and the ground-truth of new data to further filter overlapped boxes. We delete the bounding box from OM if it has an IoU greater than $\theta_{IoU}$ with a ground-truth bounding box of a new class, which are obviously wrong detection results. The remaining bounding boxes as the pseudo ground-truth are combined with the original annotations of new data for training IM.

\begin{figure}[t]
	\begin{minipage}[b]{0.45\linewidth}
		\centerline{\includegraphics[width=2.3cm]{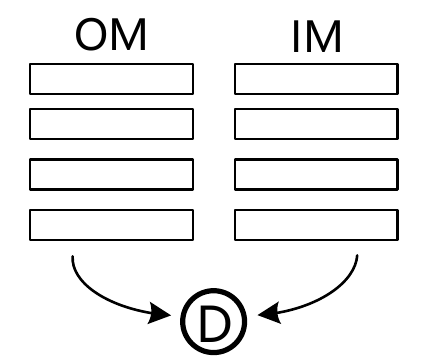}}
		\centerline{(a) General Distillation}
	\end{minipage}
	\begin{minipage}[b]{0.47\linewidth}
		\centerline{\includegraphics[width=4.1cm]{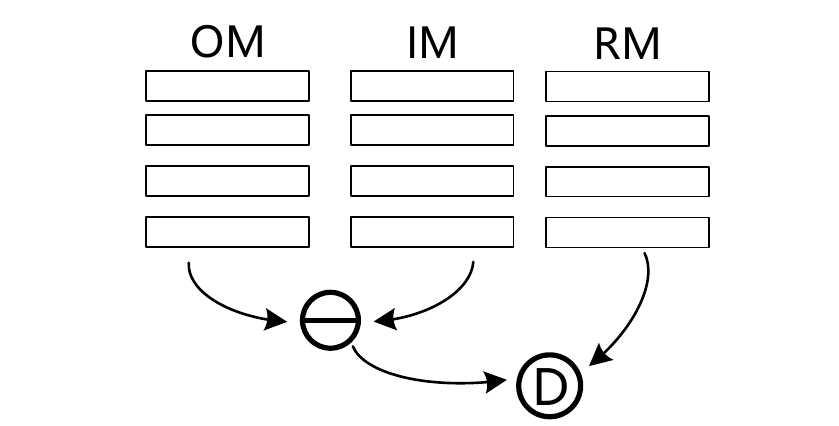}}
		\centerline{(b) Residual Distillation of Backbone}
	\end{minipage}
	\begin{minipage}[b]{1.0\linewidth}
		\centerline{\includegraphics[width=5.5cm]{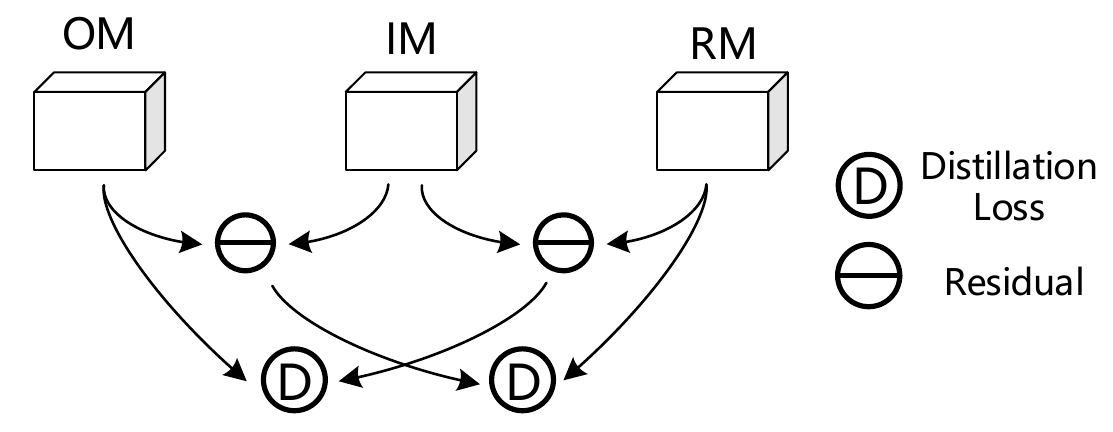}}
		\centerline{(c) Residual Distillation of Pooled Feature after RoI-Pooling}
	\end{minipage}
	\caption{Illustration of generally used distillation (a) and the residual distillation (b) (c) in this paper.}
	\label{fig:res}
\end{figure}

To preserve the learned knowledge, many methods generally train the IM to mimic the feature and outputs of OM. As shown in Figure~\ref{fig:res} (a), the general distillation calculates the difference between OM and IM, and minimizes the differences simultaneously. This constrains will restrict the capability of IM to learn the unique character of both old and new classes. Therefore, RM is designed to maintain the difference between old and new classes, which is only trained to detect new classes in the incremental learning stage. Meanwhile, RM should mimic the residual feature between OM and IM, which can be seen as the representation of new classes. The backbone of RM is initialized by the pretrained ResNet50, and the remaining parameters are initialized randomly. 

In this framework, OM and RM are jointly utilized to assist the training of IM, and only IM is utilized to get the detection results in the inference procedure. It can be noted that RM is jointly training with IM, so this method is still an end-to-end incremental learning procedure. 

\subsection{Distillation Losses}
In order to avoid catastrophic forgetting, we design a new distillation scheme to constrain the learning process of the IM, which is introduced to keep significant activations from OM, IM and RM to be similar, thus preserving the knowledge of previous seen classes and maintain the learning capability for unseen classes. The scheme consists of two type distillation losses, which are applied on different positions of the network: (1) feature space; (2) output layers of R-CNN.

For the feature space, except the general distillation between IM and OM, we also propose a two-level residual distillation, which includes base feature residual and pooled feature residual, as shown in Figure~\ref{fig:res} (b) (c). For the backbone of IM, although freezing the parameters is the best way to preserve the knowledge of previous seen classes, it will lose the ability to update for new classes, and the learning for new classes only depends on the updating of parameters in classifier and regressor, which will degrade the performance. If we directly finetune the backbone on the new data, the parameters will update towards new classes. According to the general ways, distillation should be used to maintain the feature similarity of OM and IM, so we design a new way to calculate the difference between the features of backbones, which is written as Eq.~\ref{Eq:baseatt}:   
%Therefore, we propose to apply a distillation loss on the feature space of the final activation of the backbone, which is designed to preserve similarities of feature activations between old and new models. The equation is written as Eq.~\ref{Eq:base}:
\begin{equation}
\label{Eq:baseatt}
\begin{aligned}
M_a(i,j)=\frac{1}{c}\sum^{c}_{k=1}F_a(k,i,j), \quad M_a\in{R^{h\times{w}}}\\
M_b(i,j)=\frac{1}{c}\sum^{c}_{k=1}F_b(k,i,j), \quad M_b\in{R^{h\times{w}}}\\
D_{fea}=L1Loss(\lVert M_b\cdot{M_b^T}\rVert_{2}-\lVert M_a\cdot{M_a^T}\rVert_{2})
\end{aligned}
\end{equation}
$F_a\in{R^{c\times{h}\times{w}}}$ and $F_b\in{R^{c\times{h}\times{w}}}$ represent feature maps. $c$ is the number of channels, and $h$ and $w$ are the spatial dimensions. $M_a$ and $M_b$ are the mean of $F_a$ and $F_b$ along channels, where $(i,j)$ represents a coordinate on a channel of the feature map. The difference between two feature maps is $D_{fea}$, where L1 loss is used to penalize differences in the L2-normalized outer products of $M_a$ and $M_b$. Compared to directly compute the L1 distance between two feature maps, this method considers the original 2D information of feature map as a whole not just points.

To maintain the capability of IM to preserve learned knowledge of seen classes and learn new classes simultaneously, we propose a residual distillation method ($D_{res}$), which can be divided into two parts $D_{res}^{base}$ and $D_{res}^{pool}$ representing the residual losses applied on the feature map of backbone and pooled feature after RoI pooling.

For the feature map of backbone, the residual between OM and IM is written as Eq.~\ref{Eq:residual}: 
\begin{equation}
\label{Eq:residual}
\begin{aligned}
F_{res}=F_{IM}-F_{OM}
\end{aligned}
\end{equation}

The difference is calculated between the residual $F_{res}$ and the feature map of RM $F_{RM}$, which is calculated as:

\begin{equation}
\label{Eq:baseres}
\begin{aligned}
M_{res}(i,j)=\frac{1}{c}\sum^{c}_{k=1}F_{res}(k,i,j), \quad M_{res}\in{R^{h\times{w}}}\\
M_{RM}(i,j)=\frac{1}{c}\sum^{c}_{k=1}F_{RM}(k,i,j), \quad M_{RM}\in{R^{h\times{w}}}\\
D_{res}^{base}=L1Loss(\lVert M_{res}\cdot{M_{res}^T}\rVert_{2}-\lVert M_{RM}\cdot{M_{RM}^T}\rVert_{2})
\end{aligned}
\end{equation}

To integrate the features of backbones in the triple network without increasing the model complexity in inference, a feature merge loss is proposed, which minimizes the distance between the feature map of IM and the merged feature map, where the merged feature map of OM and RM is represented by $F_{syn}$, and we keep the feature of IM close to it. Due to the residual mechanism, the sum of features from OM and RM can be regarded as the representation of incremental learning. Therefore, to better fuse these three features, we calculate the loss (Eq.~\ref{Eq:baseatt}) between the merged feature ($F_{syn}$) and IM feature.

\begin{equation}
\label{Eq:feafuse}
\begin{aligned}
F_{syn}=F_{OM}+F_{RM}
\end{aligned}
\end{equation}

%The merged feature is input into the remaining part of IM, and original features of OM and RM are input into the remaining part of them. 
After RoI-Pooling, we add another residual distillation loss for pooled features using the same RoIs from IM to assist the residual learning, where a bidirectional regularization is designed to maintain the feature discrimination in the triple-network, and L1 loss is directly applied to compute the distance between these two feature maps in instance-level.
\begin{equation}
\label{Eq:poolres}
\begin{aligned}
D_{res}^{pool}=\|P_{IM}-P_{OM},P_{RM}\|_1+\|P_{IM}-P_{RM},P_{OM}\|_1
\end{aligned}
\end{equation}
where $P$ represents the pooled feature.

For the output layers of R-CNN, the weights and biases of the classification layers can be considered as the high-level semantic representation of classes, so the parameters of classification layers corresponding to old classes are initialized from OM to maintain the representation. In the training of IM, the classifier is finetuned for learning new classes, and the classification outputs from IM is constrained to generate similar distributions to OM and RM for old and new classes respectively, ensuring to preserve the original knowledge. We calculate L2 loss between the softmax outputs of old classes and background from the classification layers of OM and IM, and the softmax outputs of new classes from the classification layers of RM and IM, which can be written as:
\begin{equation}
\label{Eq:softmax}
\begin{aligned}
C_{IM}^{old,i}=\frac{exp(y^i)}{\sum_{j=0}^{|C_A|}exp(y^{j})},& C_{OM}^{i}=\frac{exp(p_{OM}^i)}{\sum_{j=0}^{|C_A|}exp(p_{OM}^{j})}\\
C_{IM}^{new,i}=\frac{exp(y^i)}{\sum_{j=|C_A|+1}^{|C_A|+|C_B|+1}exp(y^{j})},& C_{RM}^{i}=\frac{exp(p_{RM}^i)}{\sum_{j=1}^{|C_B|}exp(p_{RM}^{j})} \\
\end{aligned}
\end{equation}
\begin{equation}
\label{Eq:cls}
\begin{aligned}
D_{cls}=\frac{1}{|C_A|+1}\sum^{|C_A|}_{k=0}& \lVert C_{IM}^{k}- C_{OM}^{k}\rVert^2 + 
\frac{1}{|C_B|}\sum^{|C_B|}_{k=1} \lVert C_{IM}^{k+|C_A|}- C_{RM}^{k}\rVert^2 \\
\end{aligned}
\end{equation}
where $C_{OM}\in{R^{|C_A|+1}}$, $C_{RM}\in{R^{|C_B|+1}}$ and $C_{IM}\in{R^{|C_A|+|C_B|+1}}$ are the classification outputs of OM, RM and IM respectively. $|C_A|$ and $|C_B|$ are the number of old and new classes respectively.

\subsection{2-Threshold Training}
For end-to-end two-stage object detectors, it is difficult to preserve the performance on old classes when directly adapt OM on new classes without using old data due to the class-sensitive RPN and R-CNN. Intuitively, the detection results of OM for new training data can provide useful information of old classes for incremental learning. Consequently, we utilize the detection results from the OM as the pseudo ground-truth to keep the ability of detecting old classes when trained on new data. Due to the lack of ground truth of old classes on new data, the wrong detection results can not be rectified. Therefore, the choice of confidence threshold is an important hyper-parameter to obtain the pseudo ground-truth, which has a great influence on the performance of IM. A high threshold may discard some potential object-like proposals of old classes, and a low threshold may contain many false negatives, which confuse the classifier of IM leading to a degraded detector.

As is well-known, for two-stage object detection methods, RPN needs to generate region proposals from complex background with a high recall, and R-CNN needs to accurately classify and regress these proposals. To better preserve the learned knowledge from old data, in the incremental training procedure, we design to train these parts of IM for different purposes, where RPN is trained to preserve the performance of RPN of OM for generating more object-like proposals for both old and new classes, and R-CNN is trained to accurately discriminate different classes. Therefore, we design a 2-threshold training strategy where a low threshold is used to select more potential proposals for training RPN and a high threshold is used to get the detection results with a high confidence for training a precise R-CNN.

Algorithm~\ref{alg:algorithm} describes the whole procedure of generating pseudo ground-truth and 2-threshold training strategy for calculating original Faster R-CNN losses. 
%The training samples of new data are fed into OM and IM simultaneously. OM generates the detection results, and then IoU is calculated between them and the ground truth of new data to discard the overlapped boxes from the generated results. The remaining results as the pseudo ground-truth are combined with the ground truth to train the incremental model. 
The losses of original Faster R-CNN consists of the losses of RPN and R-CNN, which are calculated as:

%\begin{equation}
%\label{Eq:frcnn}
%\begin{aligned}
%&l_{RPN}=loss(M_{n}.RPN((boxes_{p}>\theta_{low}) \cup gt))\\
%&l_{RCNN}=loss(M_{n}.RCNN((boxes_{p}>\theta_{high}) \cup gt))\\
%&l_{frcnn}=l_{RPN}+l_{RCNN} \quad
%\end{aligned}
%\end{equation}

\begin{equation}
\begin{aligned}
L_{RPN}^{IM}=loss(IM.RPN((boxes_{p}>\theta_{low}) \cup gt))
\end{aligned}
\label{rpn}
\end{equation}

\begin{equation}
\begin{aligned}
L_{RCNN}^{IM}=loss(IM.RCNN((boxes_{p}>\theta_{high}) \cup gt))
\end{aligned}
\label{rcnn}
\end{equation}

\begin{equation}
\begin{aligned}
L_{frcnn}^{IM}=L_{RPN}^{IM}+L_{RCNN}^{IM}
\end{aligned}
\label{frcnn}
\end{equation}

%\begin{equation}
%\begin{aligned}
%&l_{RPN}=loss(M_{i}.RPN((boxes_{p}>\theta_{low}) \cup gt))\\
%&l_{RCNN}=loss(M_{i}.RCNN((boxes_{p}>\theta_{high}) \cup gt))\\
%&l_{frcnn}=l_{RPN}+l_{RCNN}
%\end{aligned}
%\end{equation}
where $boxes_p$ is the pseudo ground-truth. The losses of RPN and R-CNN include classification and regression losses. $\theta_{low}$ and $\theta_{high}$ are two confidence thresholds, which are used to select pseudo ground-truth for training RPN and R-CNN respectively.

\begin{algorithm}[tb]
	\caption{2-Threshold Training Strategy}
	\label{alg:algorithm}
	\begin{algorithmic}[1] %[1] enables line numbers
		\Require The incremental model $IM$, old model $OM$, image $i$, ground truth $gt$, two confidence thresholds $\theta_{high}$ and $\theta_{low}$, IoU threshold $\theta_{IoU}$
		\Ensure loss $L_{frcnn}^{IM}$ 
		\State Results of old model $boxes=NMS(OM(i),\theta_{IoU})$
		\State Let pseudo ground-truth $boxes_{p}=\varnothing$%\emptyset
		\For{$box$ \textbf{in} $boxes$}
		\If{$IoU(box,gt)<\theta_{IoU}$}%$ $&$}$conf>\theta_{low}$ and
		\State $boxes_{p} = boxes_{p}\cup box$
		\EndIf
		\EndFor
		%\State Let $l_{frcnn}=0$.
		\State Compute $L_{RPN}^{IM}$ with $\theta_{low}$ (Eq.~\ref{rpn})  
		\State Compute $L_{RCNN}^{IM}$ with $\theta_{high}$ (Eq.~\ref{rcnn})  %$L_{frcnn}+=loss(IM.RPN((boxes_{p}>\theta_{low}) \cup gt))$
		%\State Compute $L_{RCNN}$ according Eq.~\ref{rcnn} %$L_{frcnn}+=loss(IM.RCNN((boxes_{p}>\theta_{high}) \cup gt))$
		\State Compute $L_{frcnn}^{IM}$ (Eq.~\ref{frcnn})
		\State \textbf{return} $L_{frcnn}^{IM}$
	\end{algorithmic}
\end{algorithm}

The overall loss $L_{all}$ used to adapt on the new classes is presented as Eq.~\ref{Eq:all}, which is the sum of the standard Faster R-CNN losses and the proposed distillation losses. The Faster R-CNN losses of IM are applied to all classes, where pseudo ground-truth of old classes and ground-truth of new classes are used for training. The Faster R-CNN losses of RM are applied to new classes, where only the ground-truth of new classes are used for training. The distillation losses are used to constrain the learning of IM.
\begin{equation}
\label{Eq:all}
\begin{aligned}
L_{all}=L_{frcnn}^{IM}+L_{frcnn}^{RM}+\lambda\times(D_{fea}+D_{res}+D_{cls})\\
%L_{frcnn}=L_{cls}^{rpn}+L_{reg}^{rpn}+L_{cls}^{rcnn}+L_{reg}^{rcnn}
\end{aligned}
\end{equation}
where $\lambda$ is a trade-off between original Faster R-CNN losses and the proposed distillation losses, we set $\lambda=1$ in all experiments.

\section{Experiments}
\begin{table*}
	\caption{Results on VOC2007 test dataset. Per-class average precision (\%) are presented under different settings when 1, 5, 10 classes are added at once.}
	\begin{center}
		\resizebox{\textwidth}{!}{
			\begin{tabular}{l|ccccccccccccccccccccc}
				\toprule
				\multicolumn{22}{c}{1}\\
				\midrule
				Method &aero & bike & bird & boat & bottle & bus & car & cat & chair & cow & table & dog & horse & mbike & person & plant & sheep & sofa & \multicolumn{1}{c|}{train} & \multicolumn{1}{c|}{tv} & mAP \\
				\midrule
				A(1-19) & 76.79&81.43 &75.85&59.2  &56.75   &81.61&84.76&84.15&51.76  &82.5 &67.23  &83.28 &83.69 &79.66  &78.21   &47.08  &73.42  &67.95 &\multicolumn{1}{c|}{77.98}  &\multicolumn{1}{c|}{-}   &73.33\\
				\midrule
				Finetune&31.80 &	24.68 &	28.27 &	25.46 &	24.59 &	43.58 &	61.38 &	35.25 &	10.60 &	35.59 &	17.47 &	22.34 &	27.46 &	20.02 &	20.01 &	16.81 &	28.11 &	11.10 &	\multicolumn{1}{c|}{28.67}& 	\multicolumn{1}{c|}{56.50} &	28.49 	\\
				~\cite{shmelkov2017incremental} &69.4 &79.3& 69.5& 57.4& 45.4& 78.4& 79.1& 80.5& 45.7& 76.3& 64.8& 77.2& 80.8& 77.5& 70.1& 42.3& 67.5& 64.4& \multicolumn{1}{c|}{76.7}& \multicolumn{1}{c|}{62.7}& 68.3\\
				~\cite{chen2019new}&\multicolumn{19}{c|}{68.30}& \multicolumn{1}{c|}{\textless60.0}& \textless68.30\\
				~\cite{li2019rilod} &69.7& 78.3& 70.2 &46.4 &59.5& 69.3& 79.7 &79.9& 52.7& 69.8 &57.4 &75.8 &69.1 &69.8& 76.4& 43.2 &68.5& 70.9& \multicolumn{1}{c|}{53.7} &\multicolumn{1}{c|}{40.4}&65.00\\
				+B(20) $D_{fea}$ $D_{res}$ $D_{cls}$ 2-th&73.65 &	81.03 &	75.17 &	60.59 &	57.69 &	80.95 &	84.65 &	85.51 &	52.11& 	80.82 &	63.67 &	83.22 &	83.89 &	80.75 &	78.04 &	47.25 &	75.04 &	67.05 &	\multicolumn{1}{c|}{79.47} &	\multicolumn{1}{c|}{51.98}& \textbf{72.13}\\ 
				\midrule
				\multicolumn{22}{c}{5}\\
				\midrule
				Method& aero & bike & bird & boat & bottle & bus & car & cat & chair & cow & table & dog & horse & mbike &\multicolumn{1}{c|}{person} & plant & sheep & sofa & train & \multicolumn{1}{c|}{tv} & mAP \\
				\midrule
				A(1-15) & 77.54&79.02&74.41&60.44&58.09&76&84.88&84.82&51.15&76&65.68&83.16&84.11&79.05&\multicolumn{1}{c|}{78.2}&-&-&-&-&\multicolumn{1}{c|}{-}&74.17\\
				\midrule
				Finetune& 54.07 &	50.34 &	47.80 	&32.71 &	21.12& 	51.57& 	71.14 &	64.62 &	19.18 &	47.98 &	47.59 &	52.77 &	61.22 &	46.08 &	\multicolumn{1}{c|}{42.46} &	37.22 &	55.63 &	56.95 &	62.99 &	\multicolumn{1}{c|}{63.31} &	49.34 \\
				~\cite{shmelkov2017incremental}&70.5& 79.2& 68.8& 59.1& 53.2 &75.4& 79.4& 78.8& 46.6& 59.4& 59.0& 75.8& 71.8 &78.6& \multicolumn{1}{c|}{69.6}& 33.7& 61.5& 63.1& 71.7& \multicolumn{1}{c|}{62.2}& 65.9\\
				+B(16-20) $D_{fea}$ $D_{res}$ $D_{cls}$ 2-th&75.56& 	81.05 &	75.76 &	58.77 &	58.11& 	77.03& 	83.90 &	84.69 &	52.77 &	75.62 &	66.25 &	81.56 &	84.37 &	78.78 &	\multicolumn{1}{c|}{76.89} &	30.83& 	65.86 &	57.98 &	72.63 &	\multicolumn{1}{c|}{55.76} &	\textbf{69.71}\\
				\midrule
				\multicolumn{22}{c}{10}\\
				\midrule
				Method& aero & bike & bird & boat & bottle & bus & car & cat & chair & \multicolumn{1}{c|}{cow} & table & dog & horse & mbike & person & plant & sheep & sofa & train & \multicolumn{1}{c|}{tv} & mAP \\
				\midrule
				A(1-10) &  90.42&90.77&90.55&90.62&86.65&87.37&90.35&89.15&87.64&\multicolumn{1}{c|}{77.78}&-&-&-&-&-&-&-&-&-&\multicolumn{1}{c|}{-}&88.13\\
				\midrule
				Finetune& 52.67 &	27.05& 	41.87 &	30.06 &	15.42 &	40.78 &	46.85 &	60.44 &	13.03 &	\multicolumn{1}{c|}{40.50} &	57.56 &	70.85 &	78.76 &	70.35 &	75.84& 	38.65 &	64.96 &	63.43 &	69.96 &	\multicolumn{1}{c|}{64.04} &	51.15 \\
				~\cite{shmelkov2017incremental} &69.9& 70.4& 69.4& 54.3& 48.0& 68.7& 78.9& 68.4& 45.5& \multicolumn{1}{c|}{58.1}& 59.7& 72.7& 73.5& 73.2& 66.3& 29.5& 63.4& 61.6& 69.3&\multicolumn{1}{c|}{62.2}& 63.1\\
				~\cite{li2019rilod} &71.7& 81.7 &66.9 &49.6 &58.0 &65.9& 84.7 &76.8&50.1 & \multicolumn{1}{c|}{69.4}& 67.0 &72.8 &77.3 &73.8 &74.9& 39.9 &68.5& 61.5 &75.5 &\multicolumn{1}{c|}{72.4}&\textbf{67.90}\\
				+B(11-20) $D_{fea}$ $D_{res}$ $D_{cls}$ 2-th&75.85 &	73.44 &	72.35 &	58.57 &	58.86 &	79.11 &	82.55 &	77.47 &44.10 &		\multicolumn{1}{c|}{73.90} &	54.20 &	73.23 &	76.15 &	72.05 &	69.86 &	30.82 &	65.05 &	56.36 &	70.99 &	\multicolumn{1}{c|}{59.24} &\underline{66.21}\\
				\midrule
				A(1-20) & 78.94& 78.94&74.87&64.61 &56.06   &81.80&84.58&84.67&52.48  &83.56&66.72  &84.60 &84.21 &78.47  &78.33   &47.93  &74.84  &69.43 &78.6 &\multicolumn{1}{c|}{73.36}&73.85\\
				\bottomrule
		\end{tabular}}
	\end{center}
	\label{table:all-detail}
\end{table*}
\begin{table}
	\caption{Results on COCO minival (first 5000 validation images). The mAP@.5 and mAP@[.5,.95] (\%) of different methods are listed when 40 classes are added at once.}
	\begin{center}
		\resizebox{0.5\linewidth}{!}{
			\begin{tabular}{l|l|l}
				\toprule
				Method & mAP@.5 &mAP@[.5, .95]\\%&old mAP@.5 &new mAP@.5
				\midrule
				A(1-40) & 32.55&16.3\\
				~\cite{shmelkov2017incremental}&37.4 &21.3\\
				+B(41-80)$D_{fea}$ $D_{res}$ $D_{cls}$ 2-th&\textbf{43.75}&\textbf{24.23}\\
				\midrule
				A(1-80) & 49.59&29.04\\
				\bottomrule
		\end{tabular}}
	\end{center}
	\label{table:coco}
\end{table}

\begin{table*}
	\caption{Results on VOC2007 test dataset. Per-class average precision (\%) are presented under different settings when 5 classes are added at once or sequentially.}
	\begin{center}
		\resizebox{\textwidth}{!}{
			\begin{tabular}{l|ccccccccccccccc|ccccc|c|c}
				\toprule
				Method & aero & bike & bird & boat & bottle & bus & car & cat & chair & cow & table & dog & horse & mbike & person & plant & sheep & sofa & train & tv & mAP & ~\cite{shmelkov2017incremental} \\
				\midrule
				A(1-15) & 77.54&79.02&74.41&60.44&58.09&76   &84.88&84.82&51.15&76   &65.68&83.16&84.11&79.05&78.2&-&-&-&-&-&74.17& \\
				\midrule
				+B(16)  &75.51 &	79.51 &	74.52 &	58.63 &	57.12 &	74.35 &	84.72 &	84.49 &	47.85 &	73.35 &	60.17 &	82.45 &	84.01 &	79.15 &	77.32 &	26.73 &	- &	- &	-& 	- &	\textbf{69.99} &67.0\\
				+B(16)(17)&74.10 &	79.87 &	73.57 &	54.91 &	57.09 &	75.65 &	84.28 &	80.96 &	46.66 &	75.86 &	61.59 &	80.58 &	83.85 &	78.62 &	76.87 &	20.20 &	60.89 &	- &	- &	- &	\textbf{68.56} &63.9\\
				+B(16)(17)(18)&72.03 &	78.36 &	73.05 &	55.35 &	57.44 &	75.99 &	84.33 &	80.62 &	47.20 &	74.21 &	60.02 &	78.97 &	83.71 &	78.56 &	73.09 &	23.54 &	30.32 &	43.08 &- &-&	\textbf{64.99}&62.5 \\
				+B(16)(17)(18)(19)&	72.10 &	77.65 &	70.16 &	52.98 &	54.61 &	74.41 &	83.59 &	79.08 &	46.34 &	73.64 &	58.43 &	77.67 &	82.33 &	76.79 &	72.91 &	20.81 &	19.17 &	51.57 &	69.28 &	- &	\textbf{63.87} &62.4 \\
				+B(16)(17)(18)(19)(20)&67.29 &	78.56 &	69.45& 	54.44 &	55.61 &	74.45 &	83.17 &	80.16 &	44.98 &	72.41 &	49.14 &	78.09 &	82.98 &	77.15 &	74.58 &	21.00 &	13.33 &	53.48 &	14.90& 	47.16 &	59.62 &\textbf{62.4} \\
				\midrule
				A(1-20) & 78.94& 78.94&74.87&64.61 &56.06   &81.80&84.58&84.67&52.48  &83.56&66.72  &84.60 &84.21 &78.47  &78.33   &47.93  &74.84  &69.43 &78.6   &73.36&73.85&\\
				\bottomrule
		\end{tabular}}
	\end{center}
	\label{table:5-sqe}
\end{table*}

\begin{table*}
	\caption{Results on VOC2007 test dataset. Average precision (\%) are presented when adding 10 classes sequentially.}
	\begin{center}
		\resizebox{0.8\textwidth}{!}{
			\begin{tabular}{l|l|cccccccccc}
				\toprule
				\multicolumn{2}{c|}{Method} &+table & +dog & +horse & +mbike & +person & +plant & +sheep & +sofa & +train & +tv \\
				\midrule
				\multirow{1}{*}{Fast RCNN}&~\cite{shmelkov2017incremental}& \underline{65.1}&	62.5&	\underline{59.9}&	\textbf{59.8}&	\textbf{59.2}&	\textbf{57.3}&	\textbf{49.1}&	\textbf{49.8}&	\textbf{48.7}&	\textbf{49}\\
				\midrule
				\multirow{2}{*}{Faster RCNN}&
				~\cite{chen2019new}            &\textbf{66.3}&	\underline{62.6}&	54.7&	50.3&	48.8&	45.5&	38.2&	36.6&	31.2&	33.5\\
				&Ours &64.46 &	\textbf{64.14} &\textbf{61.68} &\underline{55.96} &\underline{52.82} &\underline{50.04} &\underline{48.25} &\underline{44.29}& \underline{37.80} &\underline{35.39} \\
				\bottomrule
		\end{tabular}}
	\end{center}
	\label{table:10-sqe}
\end{table*}

\begin{table}
	\caption{Results on VOC2007 test dataset, when four groups are added sequentially.}
	\begin{center}
		\resizebox{0.5\linewidth}{!}{
			\begin{tabular}{l|cccc|c|c|c}
				\toprule
				Method &A& B &C& D&mAP&~\cite{shmelkov2017incremental}&~\cite{hao2019end}\\
				\midrule
				\multirow{4}{*}{OURS}
				&71.97 &	- &	- &-	&\textbf{71.97}&66.3&63.9\\	 
				&66.23	 &69.98 &	- &	-&\textbf{68.1}  &52&57.5\\
				&60.71 & 51.24 & 60 &	-&\textbf{57.32} &47&50.9\\
				&54.89 &  44.64& 39.81  &41.02  &\underline{45.09}  &39.25&\textbf{48.5}\\
				\bottomrule
		\end{tabular}}
	\end{center}
	\label{table:4-group}
\end{table}
\subsection{Settings}
\textbf{Datasets.} The proposed method is evaluated on two object detection benchmarks PASCAL VOC2007 and Microsoft COCO. VOC2007 has 20 object classes, and consists of 5K images in trainval subset and 5K images in test subset. We use the test subset for evaluation. COCO has 80 object classes, and the minival (the first 5000 images from the validation set) split is used for evaluation. We split new and old classes the same as the setting in~\cite{shmelkov2017incremental}. The experiments are conducted with different number of classes, and the classes from VOC2007 and COCO are sorted in alphabetical order. We take 19, 15 and 10 classes from VOC2007 as old classes respectively, and the remaining 1, 5, 10 classes are the corresponding new classes. For COCO, we take the first 40 classes as old classes and the remaining 40 classes as new classes. 

\noindent\textbf{Evaluation.} The evaluation metric is mean average precision (mAP) at 0.5 IoU threshold for VOC2007, and mAP across different IoU from 0.5 to 0.95 for COCO. Our method is compared with finetuning and recent related works based on two-stage object detectors%~\cite{chen2019new}~\cite{hao2019end}~\cite{li2019rilod}~\cite{shmelkov2017incremental}
. We list the results of these methods reported in their original papers, which are evaluated under the same settings with our method. The best results are in boldface, and the second-best results are underlined. 

\noindent\textbf{Implementation details.} OM is trained for 20 epochs, and the initial learning rate is set to 0.001, and decay every 5 epochs with gamma 0.1. The momentum is set to 0.9. IM is trained for 10 epochs with initial learning rate 0.0001 and decay to 0.00001 after 5 epochs. The confidence and IoU threshold for NMS are set to 0.5 and 0.3 respectively, and the IoU threshold for filtering pseudo ground-truth is also 0.3. In the following experiments, some notations are presented. A() is the results of OM, and +B() is the results of our incremental learning method, which is training on the base of A(). We define $D_{fea}$ to represent the feature distillation between OM and IM, and $D_{res}$ represents residual distillation on feature space, and $D_{cls}$ represent the joint distillation on classification layers respectively. $2-th$ represents incremental learning with 2-threshold training strategy. 

\subsection{Addition of Classes at Once}%{Addition of One Class}
In the first experiment, we evaluate the performance of our method on VOC2007 when 1, 5, 10 new classes are added at once. The results on these three setting are listed in Table~\ref{table:all-detail}, which presents the per-category average precision on VOC2007 test subset.

For the first setting in this table, we test the performance on 19 old classes and one new class (tvmonitor) from VOC2007. We train OM on the VOC2007 trainval subset with all data containing any of 19 classes (A(1-19)), and IM is trained on the data of VOC2007 trainval subset containing ``tvmonitor" (+B(20)).% The per-category average precision on VOC2007 test subset are listed in Table~\ref{table:1-detail}. 
In our experiments, the first baseline method is finetuning, and we initialize IM by the parameters of OM. Different from original finetuning which train a new classification layer from scratch for a new task, we also initialize the parameters of old classes in the classification layer of IM by those of OM to preserve the learned knowledge. However, as can be seen from the first part in Table~\ref{table:all-detail}, finetuning gets only 28.49\% mAP on all classes when old classes are in the majority, which demonstrates catastrophic forgetting can be caused by this way. It can be noted that the combination of all distillation losses with a 2-threshold training strategy (+B(20) $D_{fea}$ $D_{res}$ $D_{cls}$ 2-th) achieves the highest accuracy (72.13\%), increasing 3.83\% compared to~\cite{shmelkov2017incremental}. The mAP also increases 0.8\% compared with the Faster RCNN based method~\cite{hao2019end}. The effectiveness of our method for mitigating catastrophic forgetting is demonstrated. 

For the second setting, we choose the first 15 classes as old classes for training OM (A(1-15)), and the remaining 5 classes are used for incremental learning. As shown in the second part in Table~\ref{table:all-detail}, although the performance of finetuning is improved with the increased number of new classes, the accuracy on old classes are still lower than our method by a large margin. The mAP of our method (+B(16-20) $D_{fea}$ $D_{res}$ $D_{cls}$ 2-th) reaches 69.71\%, and increases about 3.81\% comparing with~\cite{shmelkov2017incremental}. 

Our method is also evaluated on adding more classes (10 classes) as shown in the third part in Table~\ref{table:all-detail}. OM is trained on 10 classes first, and IM learns to detect the remaining 10 new classes. The proposed method with all distillation losses(+B(11-20) $D_{fea}$ $D_{res}$ $D_{cls}$ 2-th) achieves 66.21\% mAP and increases 3.11\% compared with~\cite{shmelkov2017incremental}. We also list the results of~\cite{li2019rilod}, which reported in the original paper under the same dataset split setting. Due to the fact that~\cite{li2019rilod} keeps some exemplars of old classes, the mAP of~\cite{li2019rilod} is slightly better than our method on the 10+10 setting. However, our method exceeds it by a large margin (7.13\%) on 19+1 setting, which demonstrate the effectiveness of our method without data of old classes.

We also test the proposed method with all distillation losses on COCO, where 40 classes are old classes and the remaining 40 classes are new classes. The results are listed in Table~\ref{table:coco}. The performance outperforms~\cite{shmelkov2017incremental} by a large margin with 6.35\% improvement on mAP@0.5 and 2.93\% on mAP@[.5,.95]. It further demonstrates the effectiveness of the proposed method on lager dataset with more classes.

\subsection{Sequential Addition of Multiple Classes}
In this experiment, we evaluate the performance of our method by adding classes sequentially for incremental learning. IM is updated on the basis of the latest trained network with a new class, and the process is repeated with another new class. For example, OM is trained on 15 old classes of VOC2007 and IM is adapted to the 16th class (+B(16)), and then a new IM uses the 16-class IM to learn the 17th class (+B(16)(17)). The process continues until the 20th class (+B(16)(17)(18)(19)(20)). The results in this scenario are listed in Table~\ref{table:5-sqe}. As can be seen, our method outperforms~\cite{shmelkov2017incremental} on the learning of the 16th, 17th, 18th and 19th classes. Compared with~\cite{shmelkov2017incremental}, the mAP after adding the 16th class increases 2.99\%, and the margin reaches to 4.66\% after adding the 17th class. The accuracy also improves after adding the 18th and the 19th classes with the mAP increasing 2.49\% and 1.47\% respectively.

We also evaluate the method on adding 10 classes sequentially as shown in Table~\ref{table:10-sqe}. Compared to the same Faster RCNN based method~\cite{chen2019new}, the mAP of our method higher in almost all incremental learning steps. 
The results of Faster RCNN based methods are worse than Fast RCNN based method after many incremental learning steps, which may result from the gradually error accumulation by previous models for end-to-end object detection methods where the process of generating region proposals also needs to learn.

To further verify the sequential addition performance, we split the trainval set of VOC2007 into four groups (A, B, C, D) as the setting in~\cite{hao2019end}, and each group contains 5 classes, where all of the 20 classes are sorted alphabetically. ResNet101 is used in this experiment for fair comparison. The results are shown in Table~\ref{table:4-group}. As can be seen, our method also achieves promising results compared with other methods.%and the mAP after the third and fourth groups added outperforms the compared methods, which demonstrate the effectiveness of our method after more incremental steps. 

\subsection{Ablation Study}
%+B(20) means the model trained on ``tvmonitor" with the pseudo ground-truth generated from OM using a confidence threshold (0.5). Compared with Fast RCNN based method~\cite{shmelkov2017incremental}, the mAP increases 0.8\%, which demonstrates the end-to-end framework with learnable RPN can also obtain better accuracy on incremental object detection with the utilization of pseudo ground-truth.Then, we add the designed distillation losses sequentially to evaluate the effectiveness of these losses for improving the accuracy. As shown in Table~\ref{table:1-detail}, the mAP increases gradually, and the mAP of the combination of all distillation losses gets 71.51\%, increasing 2.41\% compared to +B(20).

To demonstrate the effectiveness of the key components, we conduct experiments to evaluate them separately when adding 1, 5, 10 classes on VOC2007 at once. As shown in Table~\ref{table:ablation}, the first row means the model trained on new classes with the pseudo ground-truth generated from OM using a confidence threshold (0.5), and the following four rows show the results when adding the designed losses and 2-threshold training strategy separately. The ``$\uparrow$" means the increased mAP of this component when compared with the first row. As listed in the table, the base feature distillation $D_{fea}$ improves 0.67\% when only add one class, which verifies the effectiveness for mitigating catastrophic forgetting. The performance of $D_{fea}$ is slightly decreased with the increasing number of new classes when $D_{fea}$ is used alone, because it is designed for preserving the performance on old classes, and it needs to cooperate with other components for improving the mAP of all classes. $D_{res}$ increases about 1.89\% on average, and the joint distillation of the final classification layer $D_{cls}$ increases about 0.25\% when used alone. 2-threshold training strategy (2-th) is also effective for boosting the performance, which increases about 0.95\% on average.
\begin{table}
	\caption{Ablation Study}
	\begin{center}
		\resizebox{0.6\linewidth}{!}{
			\begin{tabular}{cccc|ccc}
				\toprule
				\multicolumn{4}{c|}{Components} %\multirow{2}{*}{D} & \multirow{2}{*}{D} & \multirow{2}{*}{resmodel} & \multirow{2}{*}{D}\\&\multirow{2}{*}{D} \\
				&\multicolumn{3}{c}{ mAP(\%) }\\
				\midrule
				%\hline %
				$D_{fea}$&$D_{res}$&$D_{cls}$&2-th& 1 & 5 &10\\
				\midrule
				&&&&69.10&66.05&64.08\\
				$\surd$&&&&69.77($\uparrow$0.67)&66.01($\downarrow$0.03) &63.94($\downarrow$0.14)\\
				&$\surd$&&&71.81($\uparrow$2.71) &68.75($\uparrow$2.70)&64.33($\uparrow$0.25)\\
				&&$\surd$&&69.34($\uparrow$0.24)&66.08($\uparrow$0.03)&64.55($\uparrow$0.47)\\
				&&&$\surd$&69.63($\uparrow$0.52)&66.40($\uparrow$0.35)&66.05($\uparrow$1.97)\\
				\midrule
				%$\surd$&$\surd$&&&71.42($\uparrow$2.31) &68.62($\uparrow$2.57)&64.39($\uparrow$0.31)\\
				$\surd$&&&&69.77&66.01&63.94\\
				$\surd$&$\surd$&&&71.42($+1.65$)&68.62($+2.61$)&64.39($+0.45$)\\
				$\surd$&$\surd$&$\surd$&&71.51($+0.09$)&68.95($+0.33$)&64.56($+0.17$)\\
				$\surd$&$\surd$&$\surd$&$\surd$&\textbf{72.13}($+0.62$)&\textbf{69.71}($+0.76$)&\textbf{66.21}($+1.65$)\\
				\bottomrule
		\end{tabular}}
	\end{center}
	\label{table:ablation}
\end{table}

\begin{table}
	\caption{Results on alternative distillation losses.}
	\begin{center}
		\resizebox{0.5\linewidth}{!}{
		\begin{tabular}{l|l|ccc}
			\toprule
			\multicolumn{2}{l|}{ Method} &  1& 5 &10\\
			\midrule
			\multirow{2}{*}{$D_{fea}$}
			&L1 Loss& 71.14&68.17&62.88\\
			%&L2 Loss& &&\\
			&Ours&\textbf{71.51}($\uparrow$0.37) &\textbf{68.95}($\uparrow$0.78)&\textbf{64.56}($\uparrow$1.68)\\
			\midrule
			\multirow{2}{*}{$D_{cls}$}
			&old&69.31&65.75 &64.08 \\
			&Ours&\textbf{69.34}($\uparrow$0.03) &\textbf{66.08}($\uparrow$0.33) &\textbf{64.55}($\uparrow$0.47) \\
			\bottomrule
		\end{tabular}}
	\end{center}
	
	\label{table:losses}
\end{table}
\begin{figure}[t]
	\begin{minipage}[b]{0.6\linewidth}
		\centering
		\centerline{\includegraphics[width=6.5cm]{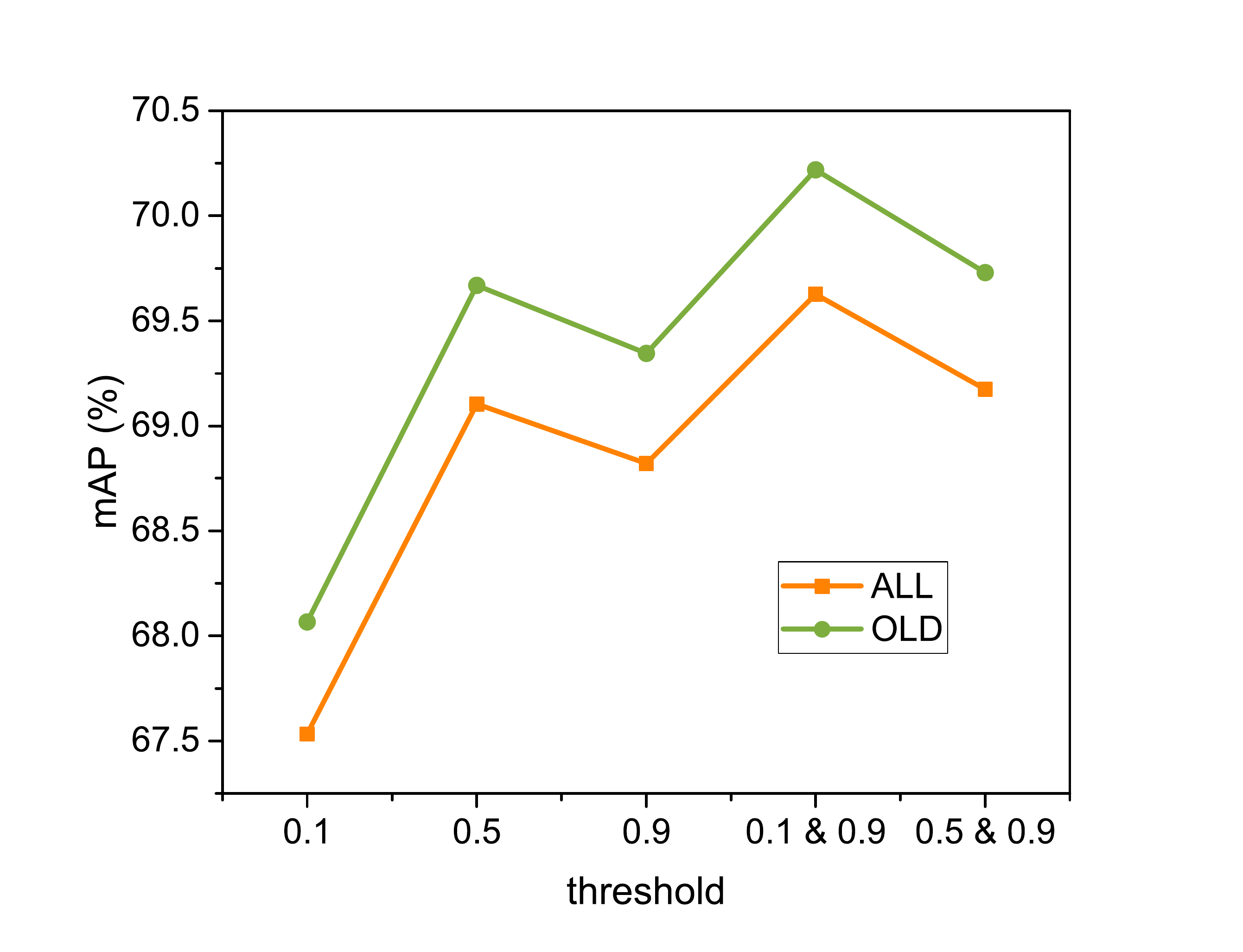}}
		%\centerline{\epsfig{figure=2th.eps,width=6.5cm}}
	\end{minipage}
	\caption{The mAP with different choices of confidence thresholds for training +B(20) network.}
	\label{fig:thresh}
\end{figure}
We also evaluate these components by adding them sequentially as shown in the last four rows of Table~\ref{table:ablation}, where ``+" represents the increased mAP of this combination compared with the last combination. As can be seen, the mAP increases gradually. The combination of $D_{fea}$ and $D_{res}$ improves about 1.57\% on the average of these three settings. When $D_{cls}$ is added, the mAP further increases about 0.2\%. The last combination of all designed components reaches the highest accuracy. This experiment further proves the validity of our method. % and the mAP of the combination of all distillation losses gets 71.51\%, increasing 2.41\% compared to +B(20).

The comparison of alternative distillation losses is shown in Table~\ref{table:losses}. The experiments are conducted on three settings (adding 1, 5, 10 new classes) respectively. To evaluate the designed L1-norm feature distillation $D_{fea}$, we replace the loss function between two 2D feature maps in $D_{fea}$ and $D_{res}$ with L1 loss, which is directly applied on the original 2D feature maps. For $D_{cls}$, the classification distillation from both OM and RM is compared with only distillation from OM. As shown, the mAP of the designed L1-norm feature distillation exceeds L1 loss about 0.94\% on average of these three settings. The joint classification distillation from both OM and RM outperforms only distillation from OM with the mAP increasing about 0.28\%. The results verify that our designs are more appropriate in this scenario.   

Figure~\ref{fig:thresh} illustrates the comparison between single-threshold training and 2-threshold training for training +B(20), where we present the results on all classes and old classes for evaluating the 2-threshold training strategy on preserving the learned knowledge of old classes. As can be seen, the 2-threshold choice with 0.1 and 0.9 can maintain the performance of old classes to a large extent and the mAP of all classes is the highest compared to other choices.

\section{Conclusion}
In this paper, we propose a triple-network based incremental object detector with a novel residual distillation scheme for learning new classes without using original training data. A frozen copy of the old model trained on old classes is used to generate pseudo ground-truth with a 2-threshold strategy and provide knowledge corresponding to old classes for training the incremental model. A residual model trained on new classes is designed to preserve the feature discrimination between old and new classes by learning the residual of the incremental model and the old model. A two-level residual distillation loss is designed for the feature of backbone and pooled feature, and a joint classification distillation is designed for the output layers. Experimental results on VOC2007 and COCO demonstrate the effectiveness of the proposed method on incrementally learning to detect objects of new classes without forgetting original learned knowledge.

%%
%% The acknowledgments section is defined using the "acks" environment
%% (and NOT an unnumbered section). This ensures the proper
%% identification of the section in the article metadata, and the
%% consistent spelling of the heading.
%\begin{acks}
%To Robert, for the bagels and explaining CMYK and color spaces.
%\end{acks}

%%
%% The next two lines define the bibliography style to be used, and
%% the bibliography file.
\bibliographystyle{ACM-Reference-Format}
\bibliography{sample-base}

%%
%% If your work has an appendix, this is the place to put it.
%\appendix

%\section{Research Methods}
%
%\subsection{Part One}
%
%Lorem ipsum dolor sit amet, consectetur adipiscing elit. Morbi
%malesuada, quam in pulvinar varius, metus nunc fermentum urna, id
%sollicitudin purus odio sit amet enim. Aliquam ullamcorper eu ipsum
%vel mollis. Curabitur quis dictum nisl. Phasellus vel semper risus, et
%lacinia dolor. Integer ultricies commodo sem nec semper.
%
%\subsection{Part Two}
%
%Etiam commodo feugiat nisl pulvinar pellentesque. Etiam auctor sodales
%ligula, non varius nibh pulvinar semper. Suspendisse nec lectus non
%ipsum convallis congue hendrerit vitae sapien. Donec at laoreet
%eros. Vivamus non purus placerat, scelerisque diam eu, cursus
%ante. Etiam aliquam tortor auctor efficitur mattis.
%
%\section{Online Resources}
%
%Nam id fermentum dui. Suspendisse sagittis tortor a nulla mollis, in
%pulvinar ex pretium. Sed interdum orci quis metus euismod, et sagittis
%enim maximus. Vestibulum gravida massa ut felis suscipit
%congue. Quisque mattis elit a risus ultrices commodo venenatis eget
%dui. Etiam sagittis eleifend elementum.
%
%Nam interdum magna at lectus dignissim, ac dignissim lorem
%rhoncus. Maecenas eu arcu ac neque placerat aliquam. Nunc pulvinar
%massa et mattis lacinia.

\end{document}